\documentclass[runningheads]{llncs}
\usepackage{graphicx}

\usepackage{tikz}
\usepackage{comment}
\usepackage{amsmath,amssymb} 
\usepackage{color}
\usepackage{appendix}
\usepackage[accsupp]{axessibility}  

\usepackage{hyperref}
\hypersetup{hypertex=true,
colorlinks=true,
linkcolor=red,
anchorcolor=blue,
citecolor=green}
\usepackage{breakcites}
\usepackage{multirow}
\usepackage{multicol}
\usepackage{mdwlist}
\usepackage{booktabs}
\usepackage{url}
\usepackage[capitalize]{cleveref}
\crefname{section}{Sec.}{Secs.}
\Crefname{section}{Section}{Sections}
\Crefname{table}{Table}{Tables}
\crefname{table}{Tab.}{Tabs.}
\usepackage[width=122mm,left=25mm,paperwidth=172mm,height=193mm,top=25mm,paperheight=243mm]{geometry}

\begin{document}
\pagestyle{headings}
\mainmatter
\def\ECCVSubNumber{4878}  
\title{Cross-Modality Knowledge Distillation Network \\ for Monocular 3D Object Detection}

\titlerunning{CMKD}

\author{Yu Hong\inst{1} \and
Hang Dai\inst{2}* \and
Yong Ding\inst{1}*}
%
\authorrunning{Y. Hong et al.}
%
\institute{Zhejiang University, Zhejiang, China \and
MBZUAI, Abu Dhabi, UAE \\
\email{yuhong\_1999@zju.edu.cn} \quad \email{hang.dai@mbzuai.ac.ae} \quad \email{dingy@vlsi.zju.edu.cn} \\
* Corresponding authors.}
\maketitle
\begin{abstract}
Leveraging LiDAR-based detectors or real LiDAR point data to guide monocular 3D detection has brought significant improvement, e.g., Pseudo-LiDAR methods.
However, the existing methods usually apply non-end-to-end training strategies and insufficiently leverage the LiDAR information, where the rich potential of the LiDAR data has not been well exploited.
In this paper, we propose the \textbf{C}ross-\textbf{M}odality \textbf{K}nowledge \textbf{D}istillation (CMKD) network for monocular 3D detection to efficiently and directly transfer the knowledge from LiDAR modality to image modality on both features and responses.
Moreover, we further extend CMKD as a semi-supervised training framework by distilling knowledge from large-scale unlabeled data and significantly boost the performance.
Until submission, CMKD ranks $1^{st}$ among the monocular 3D detectors with publications on both KITTI $test$ set and Waymo $val$ set with significant performance gains compared to previous state-of-the-art methods.
Our code will be released at \url{https://github.com/Cc-Hy/CMKD}.
\end{abstract}

\section{Introduction}

Detecting objects in 3D space is crucial to a wide range of applications, such as augmented reality, robotics and autonomous driving. The 3D detectors are to generate 3D bounding boxes with size, location, orientation and category parameters to localize and classify the detected objects, enabling the system to perceive and understand the surrounding environment.
In autonomous driving \cite{KITTI,Waymo,nuscenes}, 3D object detectors can be categorized into LiDAR point cloud based \cite{voxelrcnn,pvrcnn,pointrcnn}, stereo image based \cite{stereorcnn,rt3dstereo,disprcnn}, monocular image based \cite{MonoRCNN,monogrnet,luo2021m3dssd} and multi-modality based methods \cite{joint3d,clocs} according to the input resources. 
Compared with LiDAR sensors, monocular cameras have many unique advantages such as low price, colored information and dense perception, and monocular 3D object detection has become an active research area.
However, there exists a large performance gap between LiDAR-based 3D detectors and monocular 3D detectors due to the lack of precise 3D information in monocular images.
Thus, monocular 3D object detection is an extremely challenging task.

Recently, leveraging LiDAR-based detectors or real LiDAR point data to guide monocular 3D detection has brought significant improvement.
For example, Pseudo-LiDAR methods \cite{PL,monopl,AMOD} transform the 2D images into 3D pseudo points via depth estimation networks \cite{DORN,bts}, and use a LiDAR-based detector \cite{fpointnet,AVOD} to perform 3D detection.
Many methods \cite{PL,monopl,AMOD,CADDN,dd3d}, including most of the Pseudo-LiDAR methods, use real LiDAR point data to provide accurate 3D supervision during training, e.g., projecting the LiDAR points onto the image plane for a sparse ground truth depth map for depth supervision.

\begin{figure*}[t]
  \centering
   \includegraphics[width=0.5\linewidth]{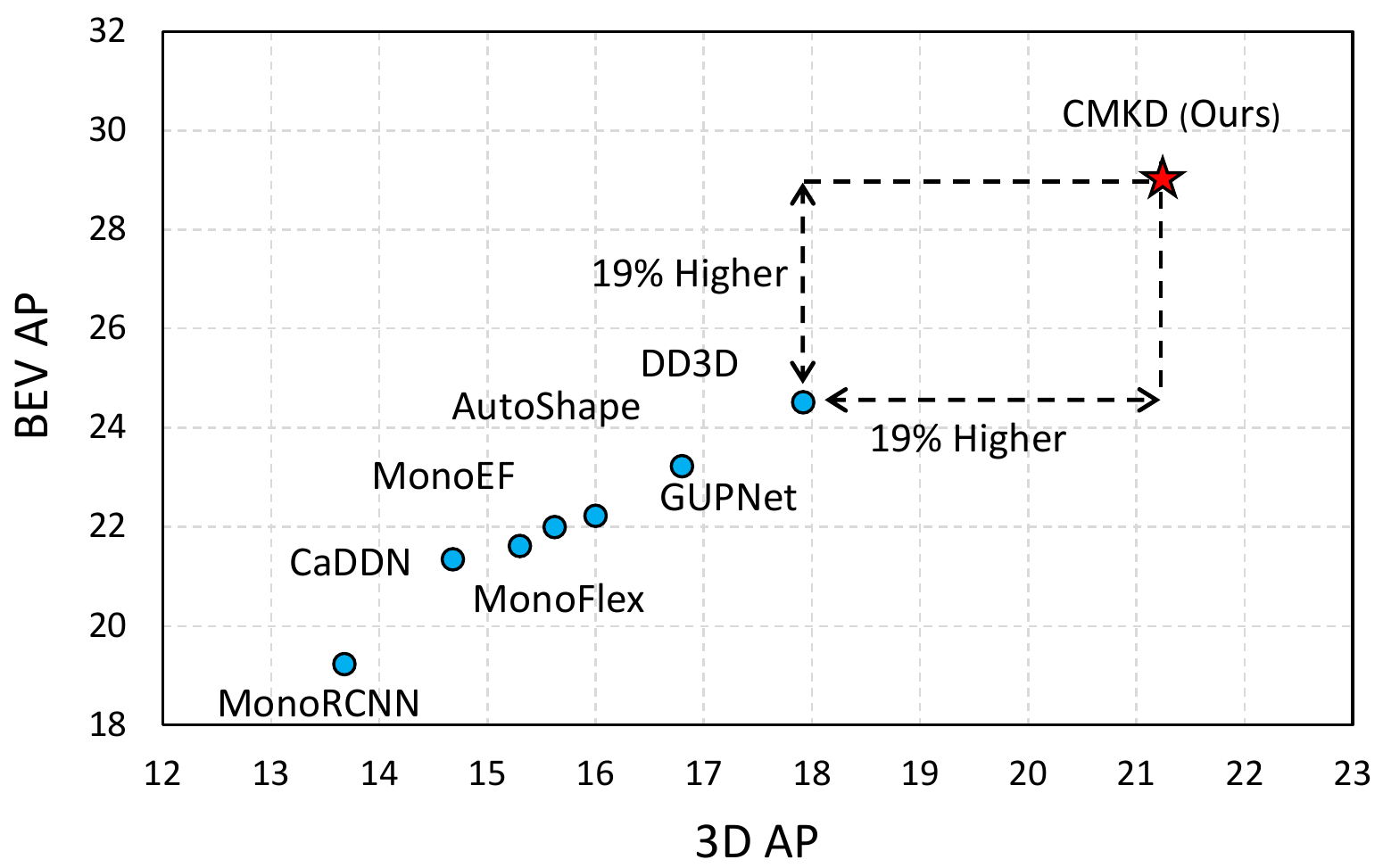}
   \caption{Comparison between top-ranking monocular 3D detectors and CMKD (Ours) on KITTI leaderboard \cite{KITTI} for Car with $3D\,AP$ and $BEV\,AP$ metrics. Higher is better.}
   \label{fig:performance}
\end{figure*}

However, there is still room for improvement in this pattern.
These methods only mimic the LiDAR data representation and extract some plain information from the LiDAR data like depth maps, but do not consider further exploiting deeper information such as high-dimensional features.
To transfer the useful knowledge from the LiDAR data more efficiently and directly, we propose a novel cross-modality knowledge distillation network to mitigate the gap between the image modality and the LiDAR modality on both features and responses.
Specifically, we use a LiDAR-based detector as the teacher model to provide the Bird’s-Eye-View (BEV) feature map which inherits accurate 3D information from LiDAR points as the feature guidance.
And we use the predictions of the teacher model with the awareness of soft label quality as the response guidance.
We then transform the knowledge from the LiDAR-based teacher model to the image-based student model in both feature and response level via distillation, thus more fully exploiting the beneficial information of the LiDAR data.

\begin{figure*}[t]
  \centering
   \includegraphics[width=0.8\linewidth]{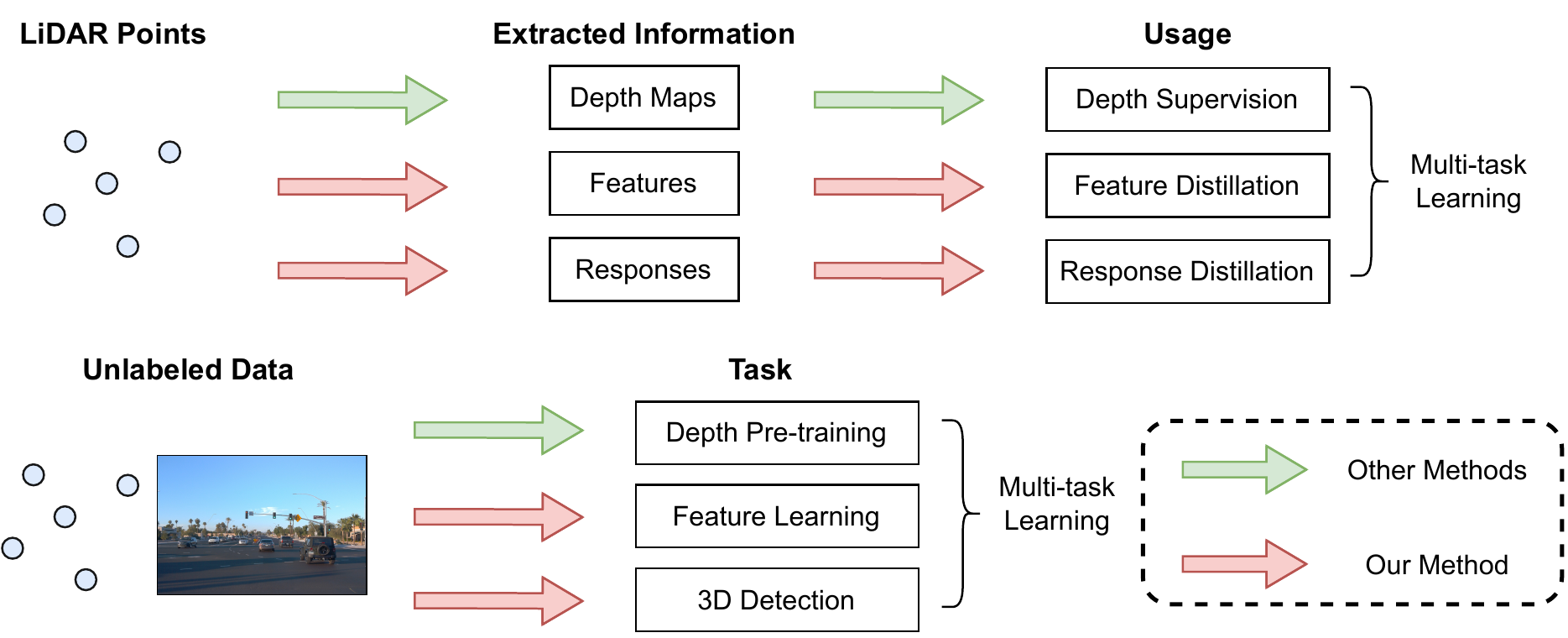}
   \caption{Comparison between other methods and CMKD (Ours). 
   For the LiDAR points, CMKD performs knowledge distillation by extracting features and responses from them, not only the depth maps. 
   For the unlabeled data, CMKD can directly use it for multi-task training including feature learning and 3D detection, not only the depth pre-training sub-task.}
   \label{fig:compare2}
\end{figure*}

Additionally, the unlabeled data, e.g., raw images and LiDAR points without ground truth 3D labels, is widely used by monocular 3D detectors \cite{PL,DA-3d,AMOD,PL++,dd3d}, but only for a sub-task like depth pre-training, and the potential of the unlabeled data has not been well exploited for the main detection task.
To this end, we further extend CMKD as a semi-supervised training framework to technically better leverage the unlabeled data.
Given a relatively small number of labeled samples to train the LiDAR-based teacher model, we can directly train CMKD on unlabeled data with the teacher model extracting beneficial information and transferring it to the student model.
Unlike the existing methods who only use the unlabeled data for depth pre-training, CMKD can directly perform the multi-task training with unlabeled data in an end-to-end manner.
Meanwhile, our semi-supervised training pipeline generalizes the application of CMKD in real-world scenes, where we only need to label a small portion of the data and can use the whole set for training, thus significantly reducing the annotation cost. 
We show the major difference between CMKD and the existing methods using LiDAR point information and unlabeled data in \cref{fig:compare2}.

We summarize our contributions in three-fold:
\textbf{i}) We propose a novel cross-modality knowledge distillation network to directly and efficiently transfer the knowledge from LiDAR modality to image modality on both features and responses,
digging deeper in cross-modality knowledge transfer and significantly improving monocular 3D detection accuracy (\cref{fig:performance}).
\textbf{ii}) We propose to distill the unlabeled data with our CMKD framework in a semi-supervised manner.
With a relatively small amount of annotated data, CMKD can be trained end-to-end on the unlabeled data, which enables it to be trained with state-of-the-art performance while significantly reducing annotation cost.
\textbf{iii}) CMKD ranks $1^{st}$ among the monocular 3D detectors with publications on KITTI $test$ set \cite{KITTI} and Waymo $val$ set \cite{Waymo} with remarkable performance gains.

\section{Related Works}

\textbf{LiDAR-based 3D Detection}\quad
LiDAR-based 3D detection \cite{pvrcnn,pointrcnn,pointnet,voxelnet,li20203d,li2021anchor,li2021voxel,li2021p2v} has been developing rapidly in recent years. 
LiDAR sensors capture precise 3D measurement information from the surroundings in the form of unordered 3D points $(x,y,z,\cdots)$,  where $x,y,z$ are the absolute 3D coordinates of each point and the others could be additional information such as reflection intensity. 
Point-based methods, e.g., PointNet \cite{pointnet}, PointNet++ \cite{pointnet++} take the raw point clouds as input, and extract point-wise features through structures like multi-layer perceptron for 3D object detection. 
Voxel-based methods, e.g., VoxelNet \cite{voxelnet}, SECOND \cite{second} extend the representation of 2D image as pixels into 3D space by dividing 3D space into voxels.
Thanks to the precise 3D information provided by point clouds, LiDAR-based methods have achieved relatively high accuracy on different 3D object detection benchmarks \cite{KITTI,Waymo,nuscenes}.

\noindent\textbf{Pseudo-LiDAR based 3D Detection}\quad
Pseudo-LiDAR based 3D detectors \cite{PL,PL++,AMOD,monopl,chen2022pseudo} benefit from both mimicking the LiDAR data representation and the accurate 3D information provided by the LiDAR data.
These methods first transform the 2D images into intermediate 3D representations like pseudo point clouds via depth estimators \cite{DORN,bts}, and then perform LiDAR-based methods on them. 
In this work, we take advantage of the LiDAR data by extracting features and responses, thus further exploiting the potential of the LiDAR data.

\noindent\textbf{Leveraging Unlabeled Data}\quad
Leveraging large-scale unlabeled data has been very popular among monocular 3D detectors especially for depth estimation pre-training. 
Pseudo-LiDAR \cite{PL} and many extension works \cite{AMOD,arewemissing,DA-3d} use an off-the-shelf depth estimator like DORN \cite{DORN} that is well-trained on the unlabeled KITTI Raw for depth estimation.
DD3D \cite{dd3d} leverages extra super-large scale unlabeled data DDAD15M for depth pre-training
which leads to significant performance improvements for monocular 3D detection.
A major improvement is that CMKD can directly use the unlabeled data to perform multi-task training in an end-to-end manner, not only the depth pre-training sub-task.

\noindent\textbf{Knowledge Distillation}\quad
The standard knowledge distillation \cite{mimicdet,distilling,likewhatyoulike,bornagain,trainingshallow,generalinstancedistillation} is performed between different models on the same modality.
Usually, a well-trained heavy teacher model is applied on the input to obtain informative representations and then supervise the features or the output logits of a simple student model, compressing the model yet maintaining high accuracy. 
In this work, we use the cross-modality knowledge distillation between the LiDAR modality and monocular image modality for monocular 3D detection.

\noindent\textbf{Difference between CMKD and Similar Methods}\quad
The general idea of knowledge distillation has been explored by some existing works, and we explain the difference.
LIGA-Stereo \cite{LIGA} focuses on the feature distillation only, and it is proposed for the stereo 3D detection task.
MonoDistill \cite{monodistill} converts the representation of LiDAR modality to image modality, while CMKD converts the representation of image modality to LiDAR modality.
LPCG \cite{LPCG} uses a LiDAR-based detector to generate pseudo labels without considering the intermediate high-dimensional features. 
Moreover, LPCG applies a one-size-fits-all method to use the soft labels, while we further take the soft label quality into account and use the quality-aware confidence scores to adaptively penalize the contribution of each soft label. 
DA-3d \cite{DA-3d} applies non-end-to-end training strategies with fixed 2D detector and depth estimator, and only the trainable feature extractor is optimized for the feature distillation. 
But the monocular detector in CMKD is fully differentiable and can be trained end-to-end with all components jointly optimized.
Overall, CMKD jointly uses feature and response distillation for the monocular 3D detection task in an end-to-end manner.
With the novel design of using totally soft guidance, CMKD can further handle large-scale unlabeled data which is easy to collect for autonomous driving cars, extending its application in real-world scenarios and boosting the performance.
Apart from the general idea of knowledge distillation, CMKD is also different in the way to perform distillation with novel explorations in each distillation module, achieving new state-of-the-art performance on KITTI and Waymo benchmarks.

\begin{figure*}[t]
  \centering
  \includegraphics[width=1\linewidth]{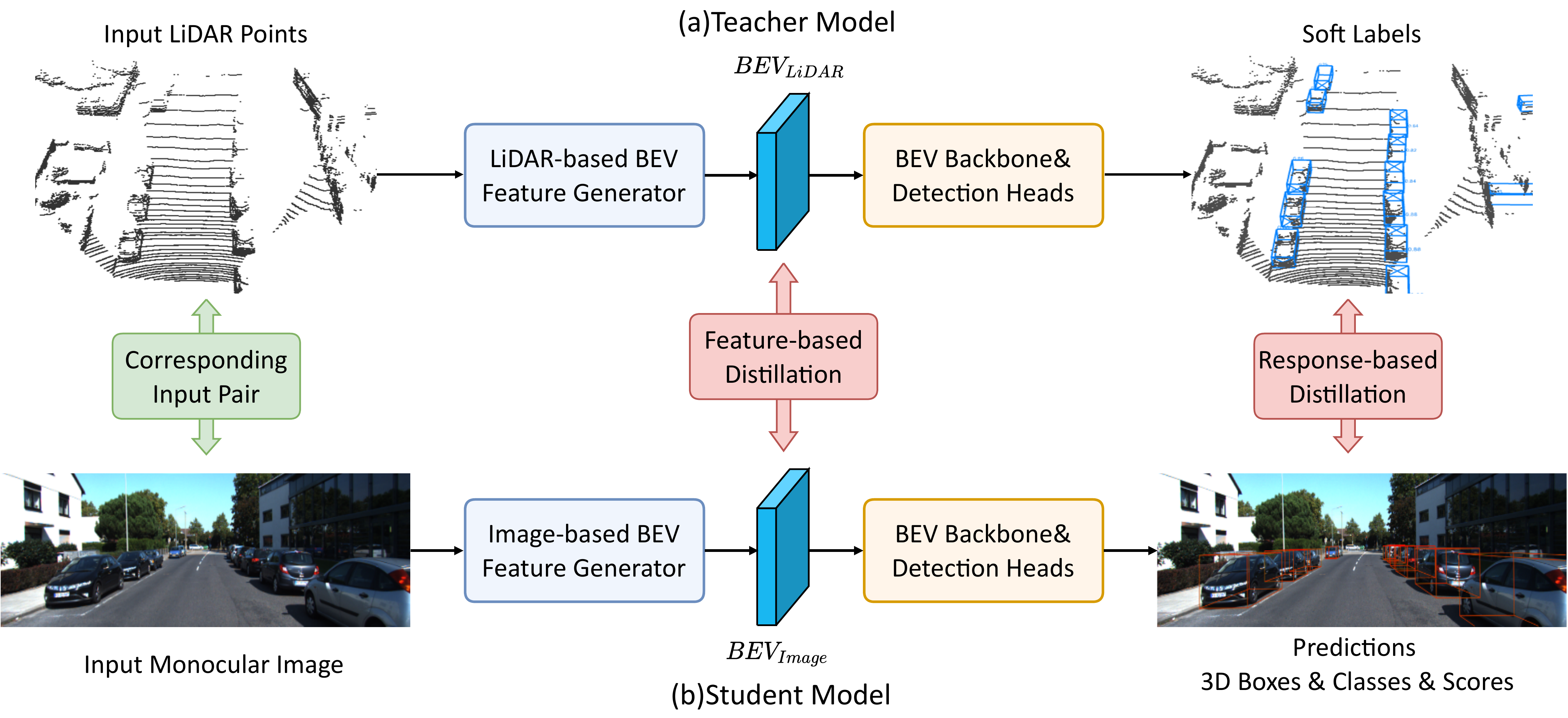}
  \caption{Overview of the cross-modality knowledge distillation (CMKD) network for monocular 3D detection. 
  (a) A pre-trained LiDAR-based 3D detector as the teacher model that extracts beneficial information from the LiDAR point data as soft guidance.
  (b) A trainable monocular 3D detector as the student model with the feature-based and response-based knowledge distillation.}
  \label{fig:overall framework}
\end{figure*}

\section{Method}

\subsection{Framework Overview}
\cref{fig:overall framework} illustrates the overview of the cross-modality knowledge distillation network for monocular 3D object detection.
The general idea is simple and straightforward.
The key is to extract the same type of feature and response representations from both input LiDAR points and input monocular images, and perform knowledge distillation between the two modalities.
Our framework includes a pre-trained LiDAR-based 3D detector as the teacher model, which extracts information from LiDAR points as soft guidance in the training stage, a trainable monocular 3D detector as the student model, and the cross-modality knowledge distillation on both features and responses.

\textbf{Training.} In the training stage, we take the monocular image and the corresponding LiDAR points as the input pair. 
The pre-trained teacher model is inferred only from input LiDAR points to provide the BEV feature maps that inherit accurate 3D information from LiDAR points as the feature guidance, and the predictions with 3D bounding boxes, object classes and their corresponding confidence scores as the response guidance. 
The student model is trainable to generate BEV feature maps and 3D object detection results from monocular images, and uses the soft guidance in both feature level and response level from the teacher model for useful knowledge transfer.

\textbf{Inference.} In the inference stage, we use the student model alone to perform 3D object detection with monocular images as input only.

\begin{figure*}[t]
  \centering
  \includegraphics[width=1\linewidth]{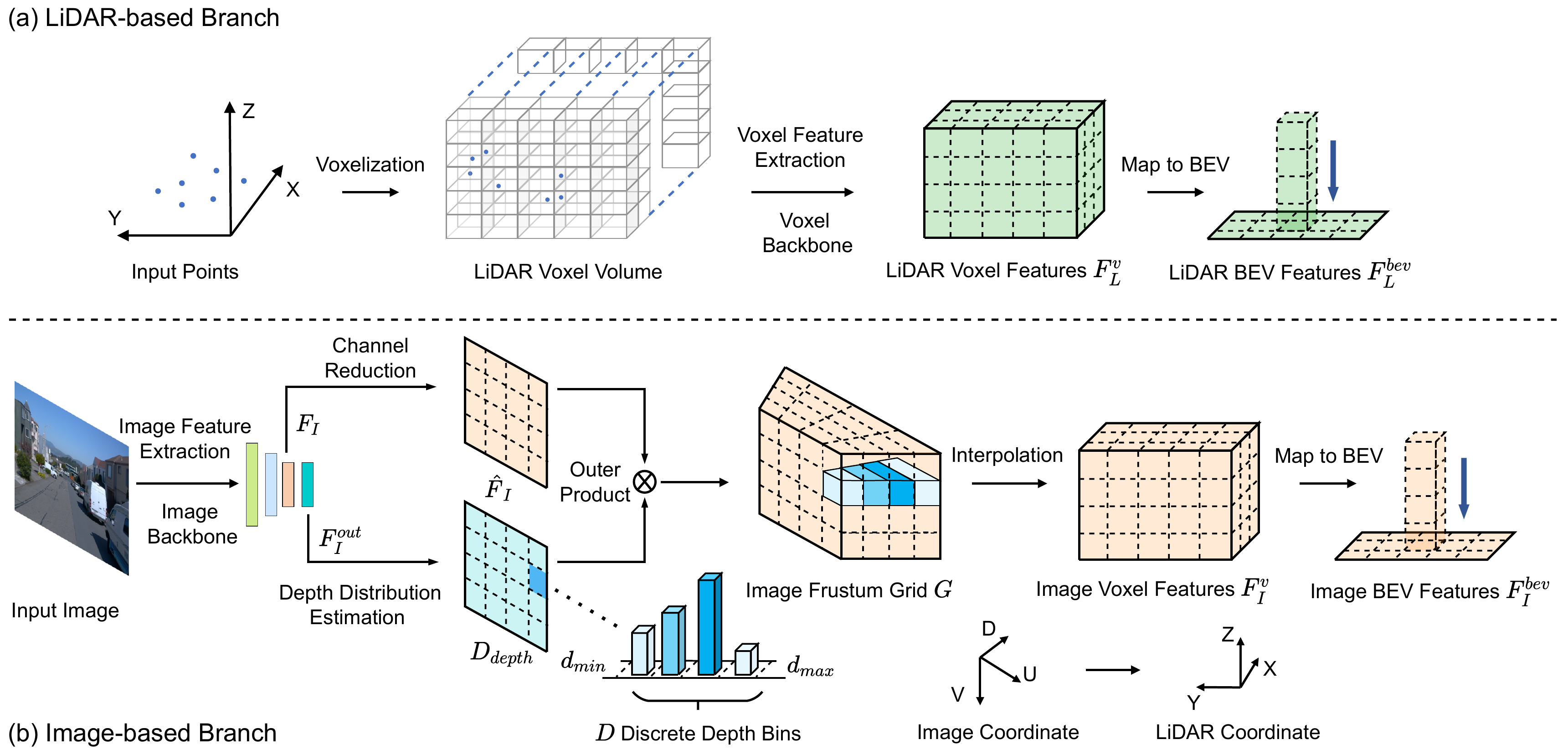}
  \caption{BEV feature map generation. (a) The LiDAR-based branch. (b) The image-based branch.}
  \label{fig:bev generation}
\end{figure*}

\subsection{BEV Feature Learning}
\textbf{LiDAR-based.} 
For the LiDAR-based model, we use SECOND \cite{second}, a simple baseline as the teacher model to extract the BEV features from LiDAR points.
The input points are first subdivided into equal 3D voxels, which are fed to a voxel backbone to extract voxel features $F^{v}_{L}\in \mathbb{R}^{X \times Y \times Z \times C}$, where $X$, $Y$, $Z$ are the width, length and height of the voxel feature volume, and $C$ is the number of feature channels. 
Then, the voxel features $F^{v}_{L}$ are collapsed to a LiDAR BEV feature map with features $F^{bev}_{L} \in \mathbb{R}^{X \times Y \times Z*C}$ by stacking the height dimension. 

\noindent\textbf{Image-based.} 
For the image-based model, we use the architecture in CaDDN \cite{CADDN} to obtain the BEV features. 
We first use an image backbone, e.g., ResNet \cite{resnet} to extract image features from the monocular image $I \in \mathbb{R}^{W\times H\times 3}$, including the intermediate image feature $F_I\in\mathbb{R}^{W_I\times H_I \times C}$ and the output image feature $F_I^{out}\in\mathbb{R}^{W_I^{out}\times H_I^{out} \times C^{out}}$.
$F_I$ goes through a channel reduction network to get $\hat{F}_{I}\in \mathbb{R}^{W_{I}\times H_{I}\times C'}$, where $C'$ is the number of the reduced feature channels.
For each position in $\hat{F}_I$, we predict its depth in a classification manner.
Specifically, the continuous depth range $[d_{min},d_{max}]$ is subdivided into $D$ discrete depth bins, and we use a depth distribution estimation head, e.g., DeepLabV3 \cite{deeplab} on $F_I^{out}$ to predict pixel-wise depth distribution $D_{depth}\in\mathbb{R}^{W_I\times H_I \times D}$ for each location in $\hat{F}_I$.
We then calculate the outer product of $\hat{F}_I\in\mathbb{R}^{W_I\times H_I \times C'}$ and $D_{depth}\in\mathbb{R}^{W_I\times H_I \times D}$ to construct a image frustum grid $G$ with features $F_G \in \mathbb{R}^{W_{I}\times H_{I}\times D\times C'}$.
The frustum volume is then converted to a cuboid volume in LiDAR coordinate via interpolation operation with known calibration parameters, and we obtain the image voxel features $F_{I}^{v} \in \mathbb{R}^{X_I \times Y_I \times Z_I \times C'}$.
The voxel features are collapsed to a BEV feature map with features $\Tilde{F}^{bev}_{I} \in \mathbb{R}^{X_I \times Y_I \times Z_I * C'}$, which then goes through a channel compression network to obtain the image BEV feature map with features $F^{bev}_{I} \in \mathbb{R}^{X_I \times Y_I \times C'}$.

We visualize the BEV feature map generation process in \cref{fig:bev generation}. 
More details can be found in SECOND \cite{second} and CaDDN \cite{CADDN}.

\subsection{Domain Adaptation via Self-Calibration}
The image BEV features $F^{bev}_{I}$ are different from LiDAR BEV features $F^{bev}_{L}$ in spatial-wise and channel-wise feature distribution due to the fact that they come from different input modalities with different backbones. 
We employ a domain adaptation (DA) module to align the feature distribution of $F^{bev}_{I}$ to that of $F^{bev}_{L}$ and enhance $F^{bev}_{I}$ at the meantime.
Specifically, we stack five Self-Calibrated Blocks \cite{scnet} after $F^{bev}_{I}$ to apply spatial-wise and channel-wise transformations:
\begin{equation}
   \hat{F}^{bev}_{I} = DA(F^{bev}_{I})
\end{equation}
where $\hat{F}^{bev}_{I}\in \mathbb{R}^{X_I \times Y_I \times C'}$ are the enhanced BEV features after the DA module.

\begin{figure}[t]
  \centering
   \includegraphics[width=0.8\linewidth]{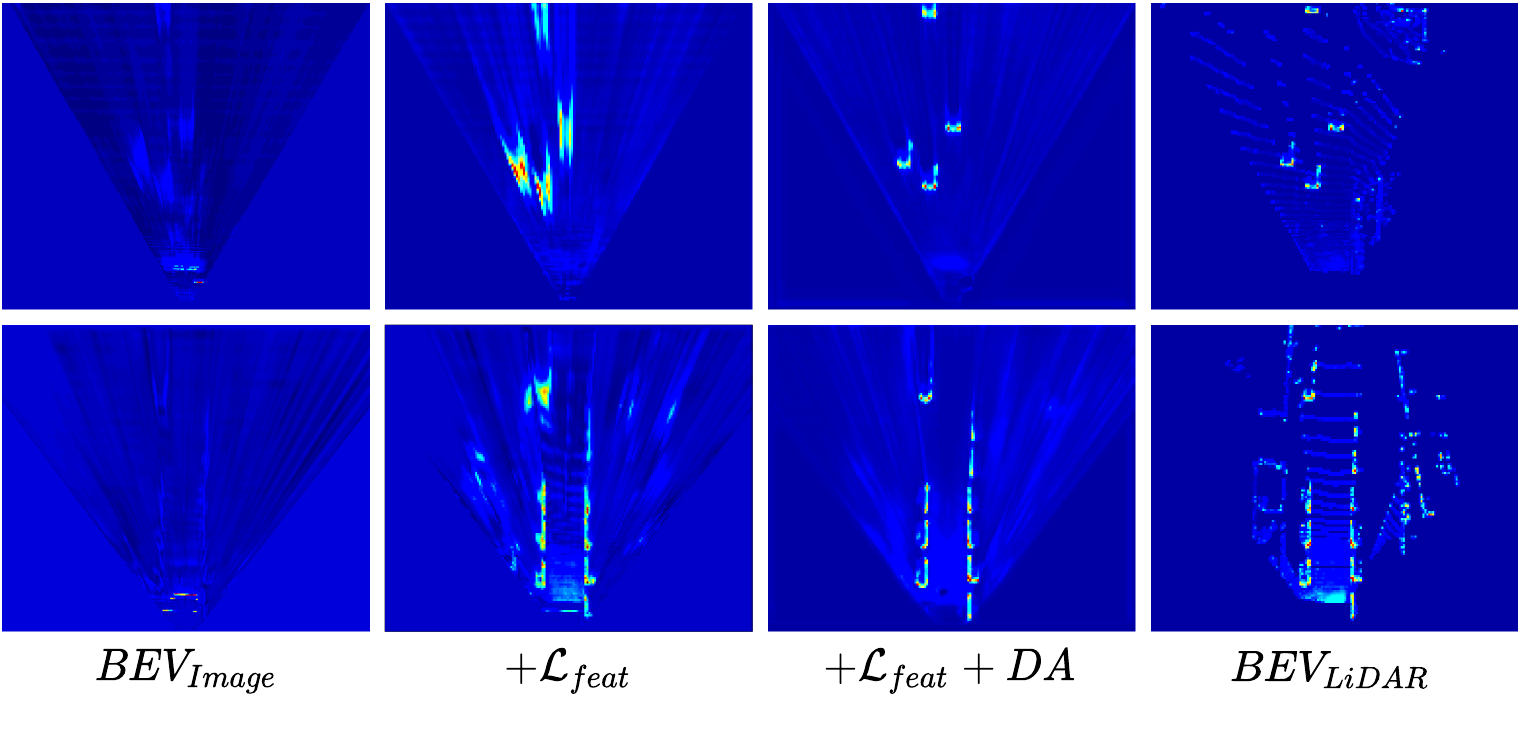}
   \caption{Illustration of BEV feature maps: the initial BEV feature map from image (1$^{st}$ column), with feature distillation loss $\mathcal{L}_{feat}$ (2$^{nd}$ column), with $\mathcal{L}_{feat}$ and DA module (3$^{rd}$ column), and the corresponding LiDAR BEV feature map (4$^{th}$ column).
}
   \label{fig:featuer distillation}
\end{figure}

\subsection{Feature-based Knowledge Distillation}
We use the BEV features $F^{bev}_{L}$ from LiDAR points as the intermediate high-dimensional feature distillation guidance for $\hat{F}^{bev}_{I}$.
We use the mean square error (MSE) to calculate the feature distillation loss:
\begin{equation}
    \mathcal{L}_{feat} = MSE(\hat{F}^{bev}_{I},F^{bev}_{L})
\end{equation}
Our monocular 3D detector benefits from the feature-based knowledge distillation due to the following aspects. 
Firstly, $F^{bev}_{L}$ contains accurate 3D information directly extracted from LiDAR points, e.g., depth and geometry.
And the feature representation of $F^{bev}_{L}$ is well-trained for 3D object detection from point clouds which is more robust to diverse scenarios such as low-light condition and weather changing. 
We can distill such patterns from $F^{bev}_{L}$ and transfer them to $\hat{F}^{bev}_{I}$.
As shown in \cref{fig:featuer distillation}, 
after feature-based knowledge distillation with the proposed DA module, the object features are highlighted and the patterns of the image BEV features are close to the LiDAR BEV features, which are the key information to detect 3D objects.
Besides, an intermediate feature guidance can ease the condition of over-fitting with high-dimensional information as the regularization term in the overall loss function \cite{fitnets,knowledgematters}.

\subsection{Response-based Knowledge Distillation}
The predictions of the teacher model are in form of $(x,y,z,h,w,l,\theta,c,s)$, where $(x,y,z)$ is the center of the 3D bounding box, $(h,w,l)$ is the size of the 3D bounding box, $\theta$ is the rotation angle, $c$ is the predicted category and $s$ is the confidence score.
And we use the predictions as the response guidance for the student model.
Compared with the hard labels, the soft labels contain more information per training sample \cite{distilling,sessd}.
Moreover, the teacher model can act as a sample filter for the training samples, e.g., samples which are very difficult to detect for the teacher model tend to be eliminated or assigned with low confidence scores, and the stable samples are assigned with high confidence scores. 

\begin{figure}[t]
    \centering
    \includegraphics[width=0.88\linewidth]{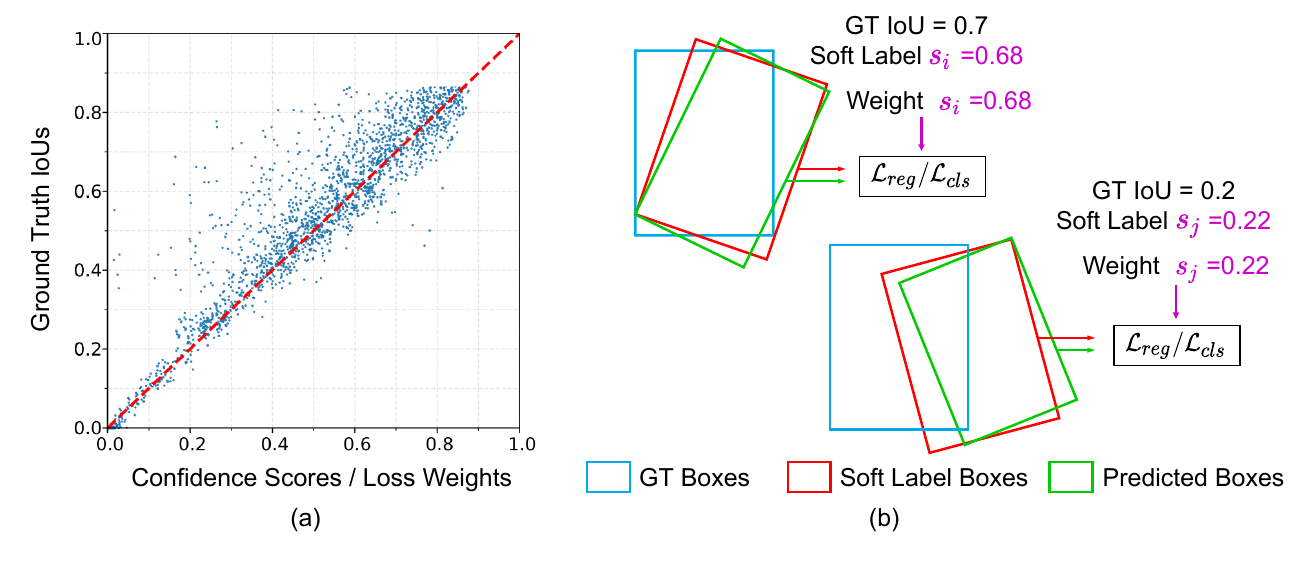}
    \caption{(a) The IoU confidence scores of soft labels are trained to be positively correlated with the ground truth IoUs.
    (b) We use the IoU confidence score of the soft label box to indicate its `quality' and weight the loss $\mathcal{L}_{reg}$/$\mathcal{L}_{cls}$ in response distillation.
    }
    \label{fig:response distillation}
\end{figure}

\textbf{Quality-aware Distillation.} The loss for response-based distillation includes the regression loss $\mathcal{L}_{reg}$ for 3D bounding boxes and the classification loss $\mathcal{L}_{cls}$ for object classes following the teacher model \cite{second}:
\begin{align}
    \mathcal{L}_{res} = \mathcal{L}_{reg} + \mathcal{L}_{cls} 
\end{align}
When pre-training the teacher model, we use the Intersection over Unions (IoUs) as the continuous quality labels with the Quality Focal Loss \cite{gfl} instead of the original one-hot labels in the classification head.
Thus, the predicted confidence scores are more IoU-aware which are used to represent the `quality' of the predictions.
For the $i$-th anchor, we use the Smooth L1 loss as the regression loss which is penalized by the IoU confidence score of the soft label:
\begin{align}
    \mathcal{L}_{reg} = Smooth\,L1(a^{soft}_i, a^{pred}_i)\times s_i         
\end{align}
where $a^{soft}_i$ and $a^{pred}_i$ are the bounding box parameters of the soft label and the prediction, and $s_i$ is the IoU confidence score of the soft label box predicted by the teacher model to indicate its `quality'. 
Similarly, we use the Quality Focal Loss (QFL) \cite{gfl} that is penalized by $s_i$ for classification:
\begin{align}
    \mathcal{L}_{cls} = QFL(C^{soft}_i, C^{pred}_i)\times s_i
\end{align}
where $C^{soft}_i$ and $C^{pred}_i$ are the classification parameters of the soft label and the prediction.
As shown in \cref{fig:response distillation}, the IoU confidence scores of the soft labels are trained to be positively correlated with their ground truth IoUs, which serve to weight the loss produced by each prediction of the student model.
Thus, our quality-aware distillation can provide more meaningful and flexible guidance.

\subsection{Loss Function}
\textbf{Teacher Model.}
We train the teacher model with the regression loss $\mathcal{L}_{reg}$ and the classification loss $\mathcal{L}_{cls}$ inherited from SECOND \cite{second} except for replacing the Focal Loss \cite{lin2017focal} with the Quality Focal Loss \cite{gfl}:
\begin{equation}
    \mathcal{L}_{teacher} = \mathcal{L}_{reg} + \mathcal{L}_{cls}
\end{equation}

\noindent\textbf{Backbone Pre-training.}
As with other methods discussed in this paper, we use the depth pre-trained backbone to make the network depth-aware, also, we initialize the backbone with the weights pre-trained on COCO \cite{lin2014microsoftcoco} before pre-training.
We inherit the depth loss from CaDDN \cite{CADDN} for backbone pre-training:
\begin{equation}
    \mathcal{L}_{pre} = \mathcal{L}_{depth}
\end{equation}

\noindent\textbf{Student Model.}
The loss function for the student model is defined as the combination of the feature-based and the response-based distillation loss:
\begin{equation}
    \mathcal{L}_{student} = \mathcal{L}_{feat} + \mathcal{L}_{res}
    \label{loss}
\end{equation}

\subsection{Extension: Distilling Unlabeled Data}
\label{extend}
After the teacher model is pre-trained with the labeled samples, every loss term in the overall loss function for the student model $\mathcal{L}_{student}$ in \cref{loss} does not use any information from manual hard labels.
Thus, we can easily and naturally extend CMKD as a semi-supervised training framework with large-scale unlabeled data that is easy to collect for autonomous driving cars. 
With the teacher model extracting beneficial information and transferring it to the student model as the soft guidance, we can use the partial labeled samples and train the model with the whole unlabeled set.
This extended ability of CMKD to handle unlabeled data significantly reduces the annotation cost and brings performance improvements, which generalizes the application of CMKD in real-world scenarios.

Note that, the utilization of unlabeled data is not new for monocular 3D detection task, especially for Pseudo-LiDAR methods.
Our contribution is to improve the utilization of unlabeled data with our cross-modality knowledge distillation network. 
The main difference is that other methods use unlabeled data only for the depth pre-training, a sub-task, but we further use it for knowledge distillation with all components of the network jointly optimized.

\section{Experiments}

\subsection{Datasets}

\textbf{KITTI 3D.}
KITTI 3D \cite{KITTI} is the most widely used benchmark for 3D object detection consisting of 7481 training images and 7518 testing images as well as the corresponding point clouds, which are denoted as KITTI $trainval$ and KITTI $test$ respectively.
The training set is commonly divided into training split with 3712 samples and validation split with 3769 samples following \cite{kittisplit}, which are denoted as KITTI\,$train$ and KITTI\,$val$ respectively.
The official evaluation metrics are 3D IoU and BEV IoU with the average precision metric, which we denote as $3D\,AP$ and $BEV\,AP$ respectively.

\noindent\textbf{KITTI Raw.}
KITTI Raw \cite{KITTIRaw} is a raw dataset with $\sim 42k$ unlabeled samples in sequence form. 
And KITTI 3D is a subset of KITTI Raw chosen with high-quality samples for 3D object detection.
Moreover, KITTI Raw is the official depth prediction training set where the training samples are commonly divided into $Eigen$ splits \cite{eigensplit}.
However, there is an overlap \cite{arewemissing,PL} between $Eigen\,train$ and KITTI $val$.
To avoid this, we use the $Eigen\,clean$ split from DD3D \cite{dd3d} that filters out KITTI $val$ from $Eigen\,train$ for the validation experiments.

\noindent\textbf{Waymo Open Dataset.}
The Waymo Open Dataset \cite{Waymo} is a more recently released dataset with 798 training sequences and 202 validation sequences which consist of about $200k$ samples in total, and we denote them as Waymo $train$ and Waymo $val$ respectively.
CaDDN \cite{CADDN} is the first monocular detector reporting the performance on Waymo $val$ set using samples from the front-camera only, and we follow the same settings for a fair comparison.
The official evaluation metrics are 3D IoU with mean average precision and mean average precision weighted by heading, which are denoted as $3D\,mAP$ and $3D\,mAPH$ respectively.

\begin{table*}[t]
  \small
  \centering
  \caption{Results for Car on KITTI $test$ set. 
    The best results are in \textbf{bold} and the second best results are \underline{underlined}.
    We present the results for two experimental setups, CMKD and CMKD*.
        CMKD is trained with the official training set KITTI $trainval$ ($\sim7.5k$) and CMKD* is trained with the unlabeled KITTI Raw ($\sim42k$).}
  \resizebox{1\textwidth}{!}{
  \begin{tabular}{l|l|cccc|cccc}
    \toprule
    \multirow{2}{*}{Methods} &\multirow{2}{*}{Reference} & \multicolumn{4}{c|}{$3D\,AP$} & \multicolumn{4}{c}{$BEV\,AP$}\\
    &   &  Easy &Moderate & Hard & Average & Easy & Moderate & Hard & Average \\
    \midrule
    M3D-PRN \cite{m3drpn} &ICCV 2019 & 14.76  & 9.71  & 7.42 &10.63  & 21.02  & 13.67  & 10.23 &14.97 \\
    AM3D \cite{AMOD} &ICCV 2019  & 16.50  & 10.74  & 9.52 & 12.25 & 25.03  & 17.32  & 14.91  & 19.08\\
    PatchNet \cite{rethinking} &ECCV 2020 & 15.68  & 11.12 & 10.17 &12.32 &22.97  & 16.86  & 14.97 & 18.27\\
    DA-3d \cite{DA-3d} &ECCV2020 &16.80 & 11.50 &8.90 & 12.40 &- &- &- &-\\
    D4LCN \cite{d4lcn} &CVPR 2020 & 16.65  & 11.72  & 9.51 & 12.63  & 22.51  & 16.02  & 12.55 &17.03\\
    Monodle \cite{monodle} &CVPR 2021 & 17.23 & 12.26  & 10.29  & 13.26 & 24.79  & 18.89  & 16.00 & 19.89 \\
    MonoRUn \cite{monorun}        &CVPR 2021  & 19.65  & 12.30  & 10.58 &14.18 & 27.94  & 17.34  & 15.24  & 20.17\\
    MonoRCNN \cite{MonoRCNN}       &ICCV 2021  & 18.36  & 12.65  & 10.03 & 13.68 & 25.48  & 18.11  & 14.10 & 19.23 \\
    PCT \cite{PCT}          &NIPS 2021 & 21.00  & 13.37  & 11.31 &15.23 & 29.65  & 19.03  & 15.92 &21.53 \\
    DFR-Net \cite{dfrnet} &ICCV 2021   & 19.40  & 13.63  & 10.35 &14.46 & 28.17  & 19.17  & 14.84 &20.73 \\
    CaDDN \cite{CADDN} &CVPR 2021   & 19.17  & 13.41  & 11.46 &14.68  & 27.94  & 18.91  & 17.19 &21.35 \\
    GUPNet \cite{gupnet}          &ICCV 2021  & 22.26  & 15.02  & 13.12 & 16.80 & 30.29  & 21.19  & 18.20 & 23.23  \\
    DD3D \cite{dd3d}     &ICCV 2021    & \underline{23.22}  & \underline{16.34}  & \underline{14.20} & \underline{17.92} & \underline{30.98}  & \underline{22.56}  & \underline{20.03} & \underline{24.52}\\
    \midrule
    \textbf{CMKD} &\multicolumn{1}{c|}{-}   & \textbf{25.09}  & \textbf{16.99}  & \textbf{15.30}  & \textbf{19.13} & \textbf{33.69}  & \textbf{23.10}  & \textbf{20.67} & \textbf{25.82} \\
    \textbf{Improvement} &\multicolumn{1}{c|}{-}  &\textbf{+1.87}  & \textbf{+0.65}  &\textbf{+1.10} &\textbf{+1.21} &\textbf{+2.71}  &\textbf{+0.54}  &\textbf{+0.64} &\textbf{+1.30} \\
    \midrule
    \textbf{CMKD*} &\multicolumn{1}{c|}{-}   & \textbf{28.55}  & \textbf{18.69}  & \textbf{16.77}  & \textbf{21.34}& \textbf{38.98}  & \textbf{25.82}  & \textbf{22.80}  & \textbf{29.20}\\
    \textbf{Improvement} &\multicolumn{1}{c|}{-} &\textbf{+5.33}  & \textbf{+2.35}  &\textbf{+2.57} &\textbf{+3.42} &\textbf{+8.00}  &\textbf{+3.26}  &\textbf{+2.77}  &\textbf{+4.68}\\
    \bottomrule
    \end{tabular}
    }
    \label{tab:car}
\end{table*}

\subsection{Experiment Settings}
\textbf{KITTI.}
We pre-train the teacher model SECOND \cite{second} on KITTI $trainval$ for 80 epochs.
For ablation studies, we train CMKD on KITTI $train$ for 80 epochs or KITTI $train$ and $Eigen\,clean$ for 30 epochs according to different experiment settings, and report the performance for Car on KITTI $val$.
The image backbone uses depth pre-training on KITTI $train$ for 40 epochs.
For comparisons on KITTI $test$, we present two experiment setups, CMKD and CMKD*.
CMKD is trained with the official training set KITTI $trainval$ ($\sim7.5k$) for 80 epochs,
and CMKD* is trained with the unlabeled KITTI Raw ($\sim42k$) for 30 epochs.
Following DD3D \cite{dd3d}, the image backbone uses depth pre-training on $eigen\,clean$ split for 10 epochs.
We report the performance for all classes on KITTI $test$. 

\noindent\textbf{Waymo.} We pre-train SECOND \cite{second} on Waymo $train$ for 10 epochs with a sampling interval 10. 
We train CMKD on Waymo $train$ for 10 epochs with a sampling interval 5 and report the performance for Vehicle on Waymo $val$. 
The input image is resized to $[960\times640]$. We do not use depth pre-training on Waymo.

\noindent\textbf{Training Skill.} 
During the training process, we adopt the following training skill to make the student model better benefit from the teacher model and make the results more stable.
Taking the 80 epochs of training on KITTI $train$ as an example, we first train the first 60 epochs with the feature distillation loss $\mathcal{L}_{feat}$ only to make the image BEV features have similar patterns to the LiDAR BEV features.
Then we load the network after the BEV features, i.e., BEV backbone and detection heads, from the LiDAR-based teacher model to the student model, and train the last 20 epochs with the complete training losses. 
In this way, the weights pre-trained in the teacher model can be utilized by the student model. 

\begin{table*}[t]
  \small
  \centering
      \caption{Results for Cyclist and Pedestrian on KITTI $test$ set. The best results are in \textbf{bold} and the second best results are \underline{underlined}.
      We present two setup results, CMKD and CMKD*.
      CMKD is trained with the official training set KITTI $trainval$ ($\sim7.5k$) and CMKD* is trained with the unlabeled KITTI Raw ($\sim42k$).}
  \resizebox{1\textwidth}{!}{
  \begin{tabular}{l|ccc|ccc}
    \toprule
    \multirow{2}{*}{Methods} & \multicolumn{3}{c|}{Cyclist\enspace\enspace$3D\,AP$ / $BEV\,AP$} & \multicolumn{3}{c}{Pedestrian\enspace\enspace$3D\,AP$ / $BEV\,AP$}\\
    {} & Easy & Moderate & Hard & Easy & Moderate & Hard \\
    \midrule
    DFR-Net \cite{dfrnet}   & 5.69 / 5.99 & 3.58 / 4.00 & 3.10 / 3.95 & 6.09 / 6.66 & 3.62 / 4.52 & 3.39 / 3.71    \\
    MonoFlex \cite{monoflex}  & 4.17 / 4.41 & 2.35 / 2.67 & 2.04 / 2.50 & 9.43 / 10.36 & 6.31 / 7.36 & 5.26 / 6.29   \\
    CaDDN \cite{CADDN}   & 7.00 / 9.67  & 3.41 / 5.38 & 3.30 / \underline{4.75} & 12.87 / 14.72  & 8.14 / 9.41  & 6.76 / 8.17 \\
    MonoPSR \cite{monopsr}   & \underline{8.37} / \underline{9.87} & \underline{4.74} / \underline{5.78} & \underline{3.68} / 4.57 & 8.37 / 9.87 & 4.74 / 5.78 & 3.68 / 4.57    \\
    GUPNet \cite{gupnet}    & 5.58 / 6.94 & 3.21 / 3.85 & 2.66 / 3.64 & \underline{14.95} / 15.62 & \underline{9.76} / 10.37 & \underline{8.41} / 8.79  \\
    DD3D \cite{dd3d}      & 2.39 / 3.20 & 1.52 / 1.99 & 1.31 / 1.79 & 13.91 / \underline{15.90} & 9.30 / \underline{10.85} & 8.05 / \underline{9.41} \\
    \midrule
    \textbf{CMKD}   & \textbf{9.60 / 12.53}  & \textbf{5.24 / 7.24}  & \textbf{4.50 / 6.21}  & \textbf{17.79 / 20.42}  & \textbf{11.69 / 13.47}  & \textbf{10.09 / 11.64}  \\
    \textbf{Improvement} &\textbf{+1.23/+2.66}  & \textbf{+0.50/+1.46}  &\textbf{+0.72/+1.46} &\textbf{+2.84/+4.52} &\textbf{+1.93/+2.62}  &\textbf{+1.68/+2.23} \\
    \midrule
    \textbf{CMKD*}   & \textbf{12.52 / 14.66}  & \textbf{6.67 / 8.15}  & \textbf{6.34 / 7.23}  & 13.94 / \textbf{16.03} & 8.79 / 10.28  & 7.42 / 8.85  \\
    \textbf{Improvement} & \textbf{+4.15/+4.79}  & \textbf{+1.93/+2.37}  &\textbf{+2.66/+2.48}  &-1.01/\textbf{+0.13}  &-0.97/-0.57  &-0.99/-0.56  \\
    \bottomrule
    \end{tabular}
    }
    \label{tab:cyc and ped}
\end{table*}

\subsection{Results on KITTI test set}
We show the results on KITTI $test$ in \cref{tab:car} and \cref{tab:cyc and ped}. 
Until submission, for all the three classes, either CMKD or CMKD* achieves new state-of-the-art results with significant improvements on KITTI $test$.
With the official KITTI $trainval$, CMKD significantly surpasses the top ranking methods. 
With additional unlabeled data from KITTI Raw and our semi-supervised training framework, CMKD* achieves further boosted performance with significant improvements for Car and Cyclist. This implies that the extension to a semi-supervised framework is efficient in distilling beneficial information from massive unlabeled data and improves the performance.
However, the performance for Pedestrian becomes worse with additional unlabeled data, and we conduct extra experiments to explore the reasons for this observation. 
This lies in the fact that the soft labels provided by the teacher model for Pedestrian are of insufficient quality, which can not provide good guidance for the student model.
Detailed experiments and discussions can be found in \cref{sec:num unlabel}.

Note that DD3D \cite{dd3d}, the top method before ours, uses large-scale extra dataset DDAD15M with $\sim15M$ samples for depth training besides KITTI, while CMKD/CMKD* uses only KITTI and surpasses DD3D by a large margin.
Also, other top methods like DD3D \cite{dd3d} or GUPNet \cite{gupnet}, works well for Car and Pedestrian but poor for Cyclist, while CMKD works well for all three object classes, which demonstrates its good generalization performance across different object classes. 

\subsection{Results on Waymo Open Dataset}
We show the results for Vehicle on Waymo $val$ in \cref{tab:vehicle waymo}. 
With fewer training samples and lower image resolution than that in M3D-RPN \cite{m3drpn} and CaDDN \cite{CADDN}, CMKD achieves significant improvements on the two difficulty levels considering different distance ranges, showing the great effectiveness.

\begin{table*}[t]
\small
\centering
\caption{
Results for Vehicle on Waymo $val$ set. The Best results are in \textbf{bold}.
We use 20\% of the training samples and half the resolution of the original image.}
\resizebox{0.7\textwidth}{!}{
\begin{tabular}{l|l|ccc|ccc}
    \toprule 
    \multirow{2}{*}{ Difficulty } & \multirow{2}{*}{ Method } & \multicolumn{3}{c|}{$3D\,mAP$} & \multicolumn{3}{c}{$3D\,mAPH$} \\
    & & Overall & 0-30m & 30-50m  & Overall & 0-30m & 30-50m \\
    \midrule 
    \multirow{4}{*}{ LEVEL\,1 } & M3D-RPN \cite{m3drpn} & 0.35 & 1.12 & 0.18  & 0.34 & 1.10 & 0.18  \\
    & CaDNN \cite{CADDN} & 5.03 & 14.54 & 1.47 & 4.99 & 14.43 & 1.45 \\
    & \textbf{CMKD} & \textbf{14.69} & \textbf{38.67} & \textbf{6.26} 
                    & \textbf{14.59} & \textbf{38.44} & \textbf{6.20}  \\ 
    & \textbf{Improvement} & \textbf{+9.66} & \textbf{+24.13} & \textbf{+4.79}  
                    & \textbf{+9.60} & \textbf{+24.01} & \textbf{+4.75} \\

    \midrule 
    \multirow{4}{*}{ LEVEL\,2 } & M3D-RPN \cite{m3drpn} & 0.33 & 1.12 & 0.18 & 0.33 & 1.10 & 0.17 \\
    & CaDNN \cite{CADDN} & 4.49 & 14.50 & 1.42 & 4.45 & 14.38 & 1.41 \\
    & \textbf{CMKD} & \textbf{12.99} & \textbf{38.17} & \textbf{5.77} 
                    & \textbf{12.90} & \textbf{37.95} & \textbf{5.71} \\ 
    & \textbf{Improvement} & \textbf{+8.50} & \textbf{+23.67} & \textbf{+4.35}  
                    & \textbf{+8.45} & \textbf{+23.57} & \textbf{+4.30} \\

    \bottomrule
\end{tabular}}
\label{tab:vehicle waymo}
\end{table*}

\begin{table}[t]
    \small
    \centering
        \caption{Effectiveness of both distillation and the extension to handle unlabeled data. 
        $Pre.$ denotes using depth pre-trained backbone.
        $Feat.$ denotes feature distillation. 
        $Res.$ denotes response distillation. 
        $Un.$ denotes distilling additional unlabeled data.}
        \resizebox{0.42\textwidth}{!}{
    \begin{tabular}{cccc|ccc}
    \toprule
    \multirow{2}{*}{$Pre.$} &\multirow{2}{*}{$Feat.$} &\multirow{2}{*}{$Res.$} &\multirow{2}{*}{$Un.$} &\multicolumn{3}{c}{$3D\,AP$} \\
    & & & & Easy & Moderate & Hard \\
    \midrule
    $\times$ & $\times$& $\times$ & $\times$ & 11.88& 8.52 & 7.40 \\
    \checkmark & $\times$& $\times$ & $\times$ & 17.60 & 13.48 & 11.81 \\
    \checkmark & $\times$& \checkmark & $\times$ & 18.81 & 14.49 & 12.16 \\
    \checkmark & \checkmark & $\times$ & $\times$ & 22.20  &15.46  &13.47 \\
    \checkmark & \checkmark & \checkmark & $\times$ &23.53 &16.33 & 14.44\\
    \checkmark & \checkmark & \checkmark & \checkmark &\textbf{30.17} &\textbf{21.54} & \textbf{19.44} \\
    \bottomrule
    \end{tabular}}
    \label{tab:abl all}
\end{table}

\subsection{Ablation Studies}
\textbf{Effectiveness of both distillation.}
As discussed earlier in this paper, existing Pseudo-LiDAR methods \cite{PL,PL++,AMOD,monopl} leverage the LiDAR data via depth pre-training,
while we further exploit the LiDAR data via knowledge distillation.
As can be seen in \cref{tab:abl all}, when using the depth pre-trained image backbone, the performance significantly improves against the baseline, indicating that the accurate depth information provided by LiDAR points is helpful for the task.
And when each of our distillation module is applied, the performance is further significantly improved, indicating that our novel utilization of the LiDAR data via distillation can more fully exploit the potential of the LiDAR data and further improve the performance of the monocular 3D detector.

\begin{table}[t]
    \small
    \centering
        \caption{Effectiveness of components in feature distillation.
    $\mathcal{L}_{feat}$ denotes the feature distillation loss. $DA$ denotes the domain adaptation module.}
    \resizebox{0.65\textwidth}{!}{
    \begin{tabular}{cc|ccc|cc|ccc}
    \toprule
    \multicolumn{5}{c|}{KITTI $train$} &\multicolumn{5}{c}{KITTI $train$ + $Eigen\,clean$}\\
    \midrule
    \multirow{2}{*}{$\mathcal{L}_{feat}$} &\multirow{2}{*}{$DA$} &\multicolumn{3}{c|}{$3D\,AP$}
    &\multirow{2}{*}{$\mathcal{L}_{feat}$} &\multirow{2}{*}{$DA$} &\multicolumn{3}{c}{$3D\,AP$}\\
    & & Easy & Moderate & Hard & & & Easy & Moderate & Hard\\
    \midrule
    $ \times$& $\times$ & 18.81 & 14.49 & 12.16 & $\times$& $\times$ & 26.07 & 19.17 & 17.45\\

    \checkmark& $ \times$ &21.72 &15.24 &12.93 & \checkmark & $\times$& 28.52 &20.74 & 18.73\\
    
    \checkmark & \checkmark &\textbf{23.53} &\textbf{16.33} & \textbf{14.44} &\checkmark & \checkmark& \textbf{30.17} &\textbf{21.54} & \textbf{19.44}\\
    \bottomrule
    \end{tabular}}
    \label{tab:abl feat}
\end{table}

\begin{table}[t]
    \small
    \centering
        \caption{Effectiveness of components in response distillation.
        $\mathcal{L}_{res}$ denotes the response distillation loss. $Conf.$ denotes the IoU-aware confidence scores of soft labels used to perform weighted supervision.}
    \resizebox{0.65\textwidth}{!}{
    \begin{tabular}{cc|ccc|cc|ccc}
    \toprule
    \multicolumn{5}{c|}{KITTI $train$} &\multicolumn{5}{c}{KITTI $train$ + $Eigen\,clean$}\\
    \midrule
    \multirow{2}{*}{$\mathcal{L}_{res}$} &\multirow{2}{*}{$Conf.$} &\multicolumn{3}{c|}{$3D\,AP$}
    &\multirow{2}{*}{$\mathcal{L}_{res}$} &\multirow{2}{*}{$Conf.$} &\multicolumn{3}{c}{$3D\,AP$}\\
    & & Easy & Moderate & Hard & & & Easy & Moderate & Hard\\
    \midrule
    $\times$& $\times$ &20.20  &13.46  &11.47 &$\times$& $\times$ & 27.24 & 19.56 & 17.67 \\
    \checkmark & $\times$& 22.78 &15.69 & 13.97 & \checkmark & $\times$& 28.16 &20.67 & 18.97\\
    \checkmark & \checkmark &\textbf{23.53} &\textbf{16.33} & \textbf{14.44} &\checkmark & \checkmark& \textbf{30.17} &\textbf{21.54} & \textbf{19.44}\\
    \bottomrule
    \end{tabular}}
    \label{tab:abl res}
\end{table}

\noindent\textbf{Effectiveness of distilling unlabeled data.}
In \cref{extend}, we introduced the improved utilization of unlabeled data in a semi-supervised manner. 
As shown in \cref{tab:abl all}, the performance of CMKD is further improved when unlabeled data is added to distillation pipeline, indicating that our method is efficient in extracting beneficial information from massive unlabeled data and improves the performance.
Specifically, we use $\sim18k$ samples for training with $\sim3.7k$ labeled and we reduce about $80\%$ annotation cost.
Also, we conducted experiments on the impact of different amounts of unlabeled data on the performance. 
Detailed experiments and discussions can be found in \cref{sec:num unlabel}.

Apart from jointly applying both distillation, we present novel designs in each distillation module, e.g., the DA module and the quality-aware supervision. We conduct experiments to show that the novel components are helpful for the task.

\noindent\textbf{Effectiveness of components in feature distillation.}
Here, the baseline is the full version of CMKD without the feature distillation loss $\mathcal{L}_{feat}$ and the DA module.
As shown in \cref{tab:abl feat}, the performance improves significantly with the two components in the feature distillation. 
As can be seen from \cref{fig:featuer distillation}, the BEV feature map shows more clear patterns with highlighted object features when $\mathcal{L}_{feat}$ is added, and avoids smearing effects with aligned BEV features when DA is added.
This shows that the components are effective in transferring the knowledge between the two modalities in the feature space.

\noindent\textbf{Effectiveness of components in response distillation.}
Here, the baseline is the full version of CMKD without the response distillation loss $\mathcal{L}_{res}$ and the quality-aware penalization weights.
As shown in \cref{tab:abl res}, the performance improves with the response distillation loss,
and achieves further improvements with the awareness of soft label quality, i.e., with the adaptive supervision.
This shows that the components are effective in transferring the knowledge between the two modalities in the response space.

\subsection{Depth Supervision vs. Feature Distillation}
Since the monocular detector in CMKD is fully differentiable and can be trained in an end-to-end manner, an option is to directly add explicit depth supervision to the model during the training stage instead of using the depth pre-training.
And this depth supervision is similar to the feature distillation in terms of motivation, where both supervision aim to make the student model learn accurate 3D information such as depth and geometry from the LiDAR data.
We conduct experiments on the two supervision for comparison.
Specifically, we remove the feature distillation module from the monocular detector in CMKD, and then add depth loss $\mathcal{L}_{depth}$ to the overall loss function. 
Also, we set up another experimental setting where we apply both depth supervision and feature distillation at the meantime.
And here, we do not use the depth pre-trained image backbone in the models using feature distillation for a thorough comparison between the two supervision.

\begin{table}[t]
    \small
    \centering
        \caption{Comparison between depth supervision and feature distillation.
        $Depth$ denotes using depth supervision. 
        $Feat.$ denotes using feature distillation.
        The results differ when using training sets with different amounts of training samples.}
    \resizebox{0.65\textwidth}{!}{
    \begin{tabular}{cc|ccc|cc|ccc}
    \toprule
    \multicolumn{5}{c|}{KITTI $train$} &\multicolumn{5}{c}{KITTI $train$ + $Eigen\,clean$}\\
    \midrule
    \multirow{2}{*}{$Depth$} &\multirow{2}{*}{$Feat.$} &\multicolumn{3}{c|}{$3D\,AP$}
    &\multirow{2}{*}{$Depth$} &\multirow{2}{*}{$Feat.$} &\multicolumn{3}{c}{$3D\,AP$}\\
    & & Easy & Moderate & Hard & & & Easy & Moderate & Hard\\
    \midrule
    \checkmark & $\times$ & \textbf{22.17} & \textbf{15.20} & \textbf{13.50} & \checkmark & $\times$ & 27.87 & 19.76 & 17.88 \\

    $ \times$ & \checkmark & 17.40 & 13.04 & 11.03 & $ \times$ & \checkmark & \textbf{31.35} & \textbf{21.32} & \textbf{19.02} \\
    
    \checkmark & \checkmark & 23.26 & 16.23 & 13.55  &\checkmark & \checkmark & 31.87 & 21.53 & 19.36 \\
    \bottomrule
    \end{tabular}}
    \label{tab:abl depth vs feat}
\end{table}

As can be seen from \cref{tab:abl depth vs feat}, when we train the models on different training sets, we get opposite results.
Specifically, when we train the models on KITTI $train$ with $\sim 3.7k$ samples which are limited, using depth supervision clearly outperforms feature distillation. And adding depth supervision on top of feature distillation can bring significant improvements.
When we train the model on KITTI $train$ and $Eigen\,clean$ with $\sim 18k$ samples, where the training data is more 
adequate, the power of feature distillation is revealed and using feature distillation clearly outperforms depth supervision.
And adding depth supervision on top of feature distillation can only bring limited improvements.

We believe this observation is within expectations.
Compared with the depth supervision, where the 3D information is supervised \textbf{explicitly}, the feature distillation supervises the depth, the geometry and the object feature representations in an \textbf{implicit} manner, where all the rich and meaningful information is fused in one layer of BEV features.
When the training samples are limited, these implicit patterns are difficult to be correctly understood by the model and the model tends to suffer from over-fitting, so in this case adding direct depth supervision which has explicit physical meanings can help the model better understand the implicit patterns represented by the BEV features and bring significant performance gains.
When the training samples become more sufficient, the rich and meaningful information provided by the feature distillation is well learned by the model, and in this case the strength of feature distillation over depth supervision is revealed.
At the meantime, the depth information, which is already contained implicitly in the feature distillation, is well learned by the model, so adding additional depth supervision can only provide limited new information and bring limited improvements to the performance.


Based on the above experiments and discussions, when the training set is small, we recommend using depth pre-training or direct depth supervision together with feature distillation.
When training samples are sufficient, feature distillation itself works well. 
Of course, we believe the potential of feature distillation can be further explored, such as designing more advanced distillation loss instead of the very simple mean square error (MSE) loss in this paper.

\subsection{Generalization Study with Different Backbones}

In this part, we conduct experiments on the generalization ability of CMKD using different student models.
For the network structure after the BEV feature map, we simply use the most basic one in 3D detection, so we mainly change the backbone of the model for comparison.
Specifically, we choose backbones with different weights and different structures and compare the performance of CMKD, including running speed, running memory and $3D\,AP$.
On the one hand, we want to show the performance of CMKD using backbones with different structures, 
on the other hand, we want to show a trade-off comparison of speed and accuracy.
The running speed and memory are tested on a single NVIDIA 3090 GPU with the batch size of 1, the $3D\,AP$ is tested on KITTI $val$.

The results are shown in \cref{tab:differentbackbone}.
We use backbones with different weights and different structures for different versions of CMKD. 
Among them, there are both heavy and deep models (running speed \textless \,10 fps), and light and shallow models (running speed \textgreater \,20 fps, which can meet the requirement for real-time applications). 
As can be seen from the table, for the Easy class, the performance gap between different models is not large, and some light-weight models perform even better than the heavy-weight ones.
For the Moderate and Hard classes, the heavy-weight models perform better than the light-weight ones, but the performance of the light-weight models is still not bad.

\begin{table}[t]
    \small
    \centering
        \caption{Generalization study of CMKD using backbones with different weights and different structures.}
        \resizebox{0.7\textwidth}{!}{
    \begin{tabular}{l|c|c|ccc}
    \toprule
   \multirow{2}{*}{Backbone} & \multirow{2}{*}{Speed (fps)} & \multirow{2}{*}{Memory (G)}  & \multicolumn{3}{c}{$3D\,AP$} \\
    & & & Easy &Moderate & Hard \\
    \midrule
    ResNet-101 \cite{resnet}& 7.5& 4.3 & {30.2} & \textbf{21.5} & \textbf{19.4} \\
    ResNet-50 \cite{resnet} & 10.1& 4.1 & 30.4 & 21.3 & 19.0\\
    EfficientNet-b5 \cite{tan2019efficientnet} & 20.8 & 2.2 & \textbf{30.8} & 20.4 & 18.5\\
    EfficientNet-b3 \cite{tan2019efficientnet} & 21.6 &2.1 & 30.7 & 20.5 & 17.9\\
    ConvNeXt-B \cite{liu2022convnext} & 23.6&2.8 & 30.6 & 20.7 & 18.5\\
    ConvNeXt-S \cite{liu2022convnext} &26.7 & 2.4& 29.7& 20.2 & 17.8\\
    MobileNet \cite{mobilenetv3} & 30.0 &2.6 & 29.8 & 20.5 & 17.8\\
    \bottomrule
    \end{tabular}}
    \label{tab:differentbackbone}
\end{table}

The above experiments, on the one hand, prove that our framework has good generalization performance and can cooperate with various backbones with different structures and weights to meet the needs of different application scenarios. 
On the other hand, it also highlights our main point of this work, that is,
what we emphasize is the idea of our cross-modality knowledge distillation (CMKD) framework, not a specific model to be used in the framework.

\subsection{Potential Limitation of CMKD: Soft Label Quality Matters}
\label{sec:limitation}
To make our work more comprehensive and complete, we proactively explore the limitations of CMKD and provide our solution. 
We notice that CMKD may have the following limitation, i.e., soft label quality matters.

Looking at the results in \cref{tab:Car}, we find that the performance of CMKD* (with $\sim42k$ training samples) on Car and Cyclist is a lot better than CMKD (with $\sim7.5k$ training samples), while the performance on Pedestrian is just the opposite, i.e., more training samples lead to worse results.
This is due to the large gap between the soft label qualities of Car, Cyclist and Pedestrian.
The typical performances of LiDAR-based detectors for Car, Cyclist and Pedestrian on KITTI leaderboard in Moderate level are around 80, 70 and 40, and the quality of predictions for Pedestrian is not at the same level as Car and Cyclist at all.
That is, the soft labels provided by the teacher model for Pedestrian themselves are of very low quality, which can not serve as good guidance for the student model.
The training of our framework on unlabeled data is under the assumption that the soft labels provided by the teacher model are of sufficient quality, which is the case for Car and Cyclist but not Pedestrian.

\begin{table*}[t]
  \small
  \centering
  \caption{Results for Car, Cyclist and Pedestrian on KITTI $test$ set. 
    CMKD is trained with the official training set KITTI $trainval$ ($\sim7.5k$) and CMKD* is trained with the unlabeled KITTI Raw ($\sim42k$).
    With unlabeled data from KITTI Raw, the performance for Car and Cyclist improves significantly, but the performance for Pedestrian instead decreases.
    }
  \resizebox{0.7\textwidth}{!}{
  \begin{tabular}{l|l|ccc|ccc}
    \toprule
    \multirow{2}{*}{Class}& \multirow{2}{*}{Methods} & \multicolumn{3}{c|}{$3D\,AP$} & \multicolumn{3}{c}{$BEV\,AP$}\\
    & & Easy &Moderate & Hard & Easy & Moderate & Hard \\
    \midrule
    \multirow{2}{*}{Car} & {CMKD}  & {25.09}  & {16.99}  & {15.30}  & {33.69}  & {23.10}  & {20.67}  \\
    & {CMKD*}  & \textbf{28.55}  & \textbf{18.69}  & \textbf{16.77} & \textbf{38.98}  & \textbf{25.82}  & \textbf{22.80} \\
    \midrule
    \multirow{2}{*}{Cyclist} & {CMKD}  & {9.60}  & {5.24}  & {4.50}  & {12.53}  & {7.24}  & {6.21}  \\
    & {CMKD*}  & \textbf{12.52}  & \textbf{6.67}  & \textbf{6.34} & \textbf{14.66}  & \textbf{8.15}  & \textbf{7.23} \\
    \midrule
    \multirow{2}{*}{Pedestrian} & {CMKD}  & \textbf{17.79}  & \textbf{11.69}  & \textbf{10.09}  & \textbf{20.42}  & \textbf{13.47}  & \textbf{11.64}  \\
    & {CMKD*}  & {13.94}  & {8.79}  & {7.42} & {16.03}  & {10.28}  & {8.85} \\
    
    \bottomrule
    \end{tabular}
    }
    \label{tab:Car}
\end{table*}

\begin{table}[t]
    \centering
    \small
    \caption{Comparison between hard labels and soft labels used in $\mathcal{L}_{res}$.
    Among them, the soft labels for Car are of sufficient quality, and the soft labels for Pedestrian are of insufficient quality. 
    We choose the early-stopped teacher model epochs to simulate the soft label quality that can be provided on unlabeled data.}
      \resizebox{0.45\textwidth}{!}{
\begin{tabular}{l|ccc}
\toprule
\multirow{2}{*}{ Settings } & \multicolumn{3}{c}{ $3D\,AP$ } \\
& Easy & Moderate & Hard \\
\midrule
\multicolumn{4}{c}{ Car, Sufficient Quality Soft Labels } \\
\midrule
Hard Labels & 23.20  & 15.78 & 13.77 \\
Soft Labels & \textbf{23.85} & \textbf{16.22} & \textbf{14.33} \\
\midrule
\multicolumn{4}{c}{ Pedestrian, Insufficient Quality Soft Labels } \\
\midrule
Hard Labels & \textbf{12.24} & \textbf{8.65} & \textbf{6.82} \\
Soft Labels & 4.57 & 3.20 & 2.47 \\
\bottomrule
\end{tabular}}
    \label{tab:hard vs soft}
\end{table}

To verify the above discussion, we conduct additional experiments.
Specifically, we choose two categories, Car and Pedestrian, and use hard label and soft label from KITTI $train$ to supervise them respectively and report the performance on KITTI $val$ for comparative experiments.
For soft labels, we choose the early-stopped teacher model epochs whose performance on KITTI $val$ are close to the typical one on the KITTI $test$ set ($3D\,AP\approx 80\%$ for Car and $3D\,AP\approx 40\%$ for Pedestrian), in order to simulate the soft label quality provided by the teacher model on unlabeled data.

As can be seen from \cref{tab:hard vs soft}, for Car, sufficient quality soft labels can provide useful information, and the results using soft labels are better than using hard labels.
But for Pedestrian, insufficient quality soft labels can not provide effective guidance, so the results are far worse than using hard labels.
When we train CMKD on unlabeled data, the teacher model can extract beneficial information for Car from the massive unlabeled data and transfer it to the student model, thereby boosting the performance of the student model.
But for Pedestrian, the soft labels provided by the teacher model themselves are of insufficient quality, which can not serve as good guidance for the student model, and on the contrary reduce the performance of the student model. 
It is for this reason that the results in \cref{tab:Car} appear.
Based on the above experiments and discussions, when the quality of the soft label is bad, the solution we provide is to change soft labels to hard labels in the loss term $\mathcal{L}_{res}$ without changing the overall framework.

\subsection{Impact of Different Amounts of Unlabeled Data}
\label{sec:num unlabel}
In this section, we conduct experiments to explore the impact of different amounts of unlabeled data on the performance. Here, the baseline is CMKD trained on KITTI $train$ with $\sim3.7k$ samples, and we gradually add unlabeled samples from $Eigen\,clean$ split to the training set. We calculate the mean $3D\,AP$ and $BEV\,AP$ for Car on KITTI $val$.

\begin{figure*}[t]
  \centering
   \includegraphics[width=1\linewidth]{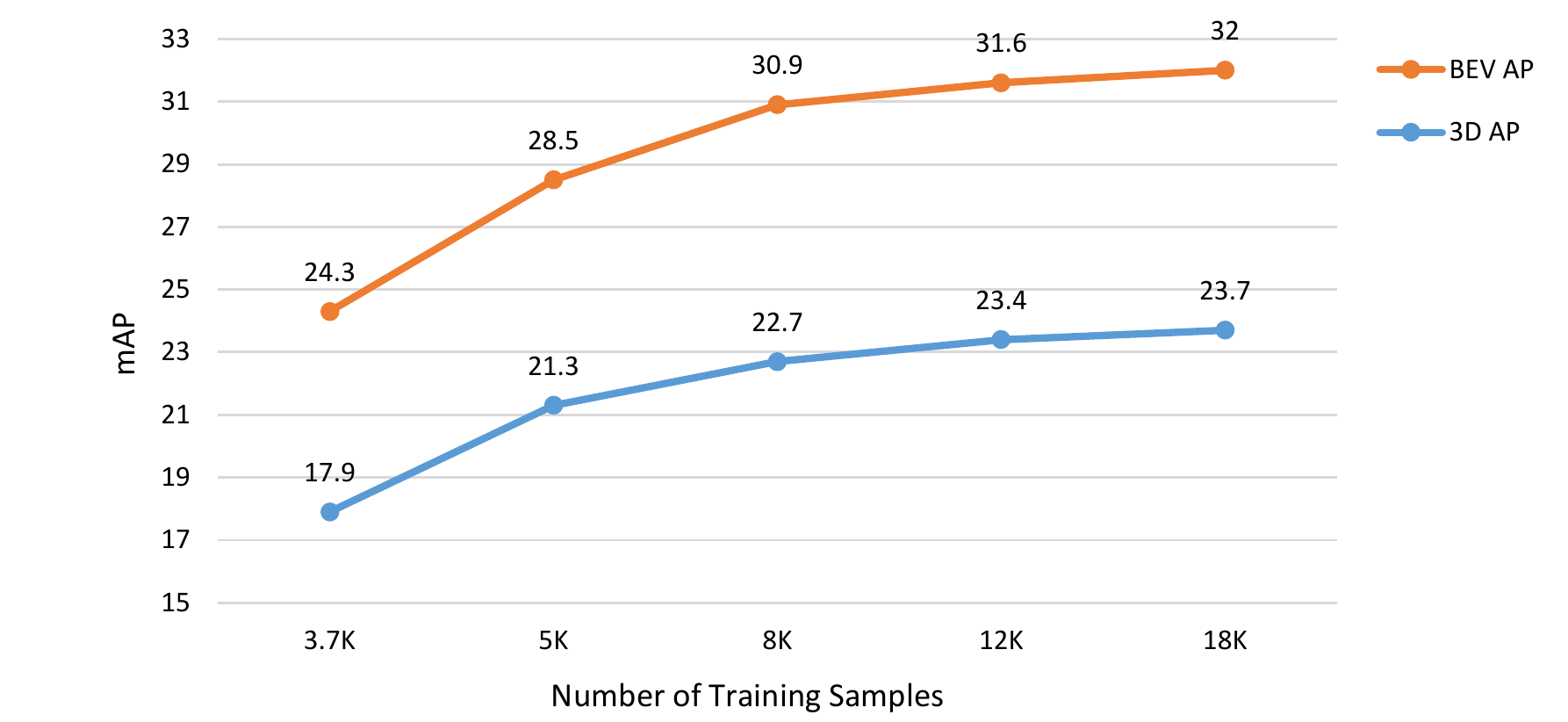}
   \caption{Impact of different amounts of unlabeled data to the performance. We use CMKD trained on KITTI $train$ with $\sim3.7k$ samples as the baseline and gradually add unlabeled samples from $Eigen\,clean$ split to the training set. We calculate the mean $3D\,AP$ and $BEV\,AP$ for Car on KITTI $val$.}
   \label{fig:unlabel}
   \vspace{+5mm}
\end{figure*}

As can be seen from \cref{fig:unlabel}, the performance of CMKD improves as the number of unlabeled samples increases. 
Specifically, when the training samples are limited (e.g., there are only $\sim3.7k$ samples on KITTI $train$, which are very few to well train a deep network like CMKD), a small number of unlabeled samples ($\sim1.3k$) can bring significant performance gains.
When the number of unlabeled samples becomes larger (+$4.3k$, +$8.3k$, +$14.3k$ respectively), the magnitude of the performance improvement tends to moderate.
And this is consistent with the trend of performance gains from pre-training with additional unlabeled data in other tasks, e.g., image classification task on ImageNet \cite{imagenet}.

Note that, here, the amount of additional unlabeled data and the information it can provide is not linear. 
As mentioned before, KITTI 3D is a sub-set of KITTI Raw and KITTI Raw is in continuous sequence form, so there are a large number of similar, repeated samples which can only provide limited new information.
Moreover, KITTI Raw is a massive unlabeled dataset which also contains many low-quality samples, e.g., with only repetitive and noisy background information, and these low-quality samples may in turn degrade the performance of the model.
However, one of our starting points of this work is that \textbf{end-to-end training can be performed directly on massive unlabeled data} to greatly reduce the cost of annotation and other pre-processing steps. 
Therefore, we do not filter these unlabeled samples, but directly use all of them for training, which is exactly the motivation of the proposed semi-supervised training method.

\subsection{Qualitative Results}

\begin{figure*}[t]
  \centering
  \includegraphics[width=1\linewidth]{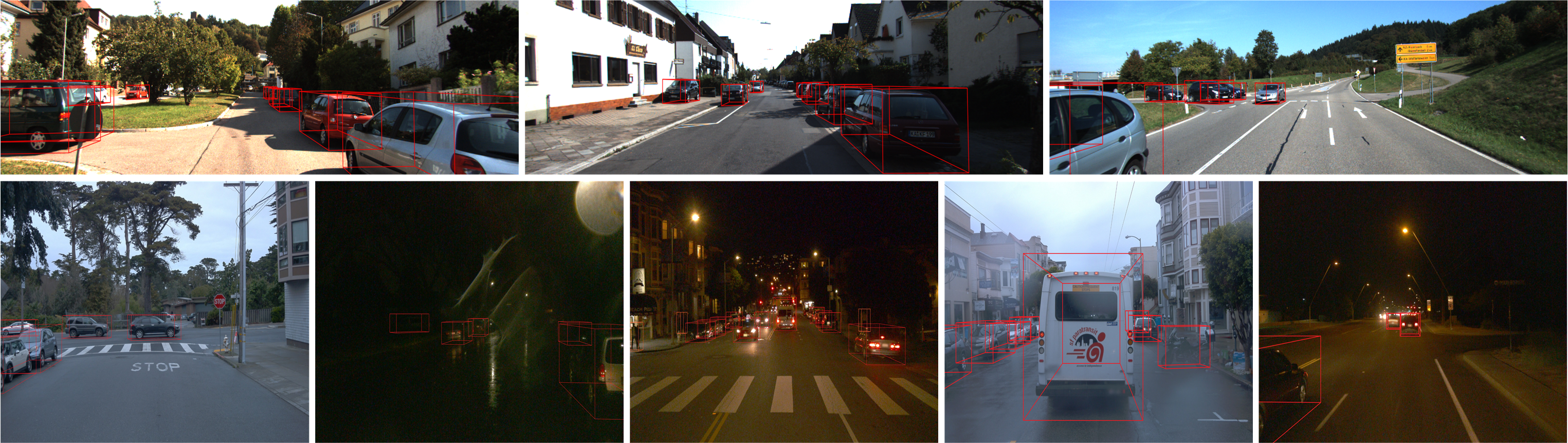}
  \caption{Qualitative results on KITTI $test$ (top line) and Waymo $val$ (bottom line). None of the samples were seen during training.}
  \label{fig:qualitive results}
\end{figure*}

We visualize some detection results from KITTI and Waymo in \cref{fig:qualitive results}.
As can be seen from the figure, the scenes from KITTI are with daytime and clear weather, while the scenes from Waymo are more complicated with light conditions and weather changes.
And CMKD works well in both datasets.

\newpage
\section{Conclusion}
In this work, we propose the cross-modality knowledge distillation (CMKD) network to directly and efficiently transfer the knowledge from LiDAR modality to image modality on both features and responses, and significantly improve monocular 3D detection accuracy.
Moreover, we extend CMKD as a semi-supervised training framework to distill useful knowledge from large-scale unlabeled data, further boosting the performance while reducing the annotation cost.
CMKD achieves new state-of-the-art performance on both KITTI and Waymo benchmarks for monocular 3D object detection with significant performance gains compared to other methods, which shows its great effectiveness.

\noindent\textbf{Broader Impact.} 
Our CMKD framework opens up a new perspective in monocular 3D detection.
We believe the effective distillation of unlabeled data demonstrates the potential of CMKD to generalize its application in real-world scenarios, where the unlabeled data is easy to collect for autonomous driving cars.

\section*{Acknowledgement}
This work was supported by the National Key Research and Development Program of China (2018YFE0183900) and YUNJI Technology Co. Ltd.

%
%

\bibliographystyle{splncs04}
\bibliography{egbib}

\begin{thebibliography}{10}
\providecommand{\url}[1]{\texttt{#1}}
\providecommand{\urlprefix}{URL }
\providecommand{\doi}[1]{https://doi.org/#1}

\bibitem{m3drpn}
Brazil, G., Liu, X.: M3d-rpn: Monocular 3d region proposal network for object
  detection. In: ICCV (2019)

\bibitem{nuscenes}
Caesar, H., Bankiti, V., Lang, A.H., et~al.: nuscenes: A multimodal dataset for
  autonomous driving. In: CVPR (2020)

\bibitem{monorun}
Chen, H., Huang, Y., Tian, W., et~al.: Monorun: Monocular 3d object detection
  by reconstruction and uncertainty propagation. In: CVPR (2021)

\bibitem{deeplab}
Chen, L., Papandreou, G., Schroff, F., et~al.: Rethinking atrous convolution
  for semantic image segmentation. CoRR  \textbf{abs/1706.05587} (2017)

\bibitem{kittisplit}
Chen, X., Kundu, K., Zhu, Y., et~al.: 3d object proposals for accurate object
  class detection. In: NIPS (2015)

\bibitem{chen2022pseudo}
Chen, Y.N., Dai, H., Ding, Y.: Pseudo-stereo for monocular 3d object detection
  in autonomous driving. In: Proceedings of the IEEE/CVF Conference on Computer
  Vision and Pattern Recognition. pp. 887--897 (2022)

\bibitem{monodistill}
Chong, Z., Ma, X., Zhang, H., Yue, Y., Li, H., Wang, Z., Ouyang, W.:
  Monodistill: Learning spatial features for monocular 3d object detection
  (2022)

\bibitem{generalinstancedistillation}
Dai, X., Jiang, Z., Wu, Z., et~al.: General instance distillation for object
  detection. In: CVPR (2021)

\bibitem{imagenet}
Deng, J., Dong, W., Socher, R., et~al.: {ImageNet: A Large-Scale Hierarchical
  Image Database}. In: CVPR (2009)

\bibitem{voxelrcnn}
Deng, J., Shi, S., Li, P., et~al.: Voxel r-cnn: Towards high performance
  voxel-based 3d object detection. In: AAAI (2021)

\bibitem{d4lcn}
Ding, M., Huo, Y., Yi, H., et~al.: Learning depth-guided convolutions for
  monocular 3d object detection. CVPR  (2020)

\bibitem{eigensplit}
Eigen, D., Puhrsch, C., Fergus, R.: Depth map prediction from a single image
  using a multi-scale deep network. In: NIPS (2014)

\bibitem{Waymo}
Ettinger, S., Cheng, S., Caine, B., et~al.: Large scale interactive motion
  forecasting for autonomous driving : The waymo open motion dataset. CoRR
  \textbf{abs/2104.10133} (2021)

\bibitem{DORN}
Fu, H., Gong, M., Wang, C., others.: {Deep Ordinal Regression Network for
  Monocular Depth Estimation}. In: CVPR (2018)

\bibitem{bornagain}
Furlanello, T., Lipton, Z.C., Tschannen, M., et~al.: Born-again neural
  networks. In: Proceedings of International Conference on Machine Learning
  (ICML) (2018)

\bibitem{KITTIRaw}
Geiger, A., Lenz, P., Stiller, C., Urtasun, R.: Vision meets robotics: The
  kitti dataset. International Journal of Robotics Research (IJRR)  (2013)

\bibitem{KITTI}
Geiger, A., Lenz, P., Urtasun, R.: Are we ready for autonomous driving? the
  kitti vision benchmark suite. In: CVPR (2012)

\bibitem{knowledgematters}
G{\"{u}}l{\c{c}}ehre, {\c{C}}., Bengio, Y.: Knowledge matters: Importance of
  prior information for optimization. In: ICLR (2013)

\bibitem{LIGA}
Guo, X., Shi, S., et~al.: Liga:learning lidar geometry aware representations
  for stereo-based 3d detector. In: ICCV (2021)

\bibitem{resnet}
He, K., Zhang, X., Ren, S., et~al.: Deep residual learning for image
  recognition. In: CVPR (2016)

\bibitem{distilling}
Hinton, G.E., Vinyals, O., Dean, J.: Distilling the knowledge in a neural
  network. CoRR  \textbf{abs/1503.02531} (2015)

\bibitem{mobilenetv3}
Howard, A.G., Sandler, M., Chu, G., et~al.: Searching for mobilenetv3. ICCV
  (2019)

\bibitem{likewhatyoulike}
Huang, Z., Wang, N.: Like what you like: Knowledge distill via neuron
  selectivity transfer. CoRR  \textbf{abs/1707.01219} (2017)

\bibitem{joint3d}
Ku, J., Mozifian, M., Lee, J., Harakeh, A., Waslander, S.L.: Joint 3d proposal
  generation and object detection from view aggregation. In: {IROS} (2018)

\bibitem{AVOD}
Ku, J., Mozifian, M., Lee, J., et~al.: Joint 3d proposal generation and object
  detection from view aggregation. In: IEEE/RSJ International Conference on
  Intelligent Robots and Systems (IROS) (2018)

\bibitem{monopsr}
Ku*, J., Pon*, A.D., Waslander, S.L.: Monocular 3d object detection leveraging
  accurate proposals and shape reconstruction. In: CVPR (2019)

\bibitem{rt3dstereo}
Königshof, H., Salscheider, N.O., Stiller, C.: {Realtime 3D Object Detection
  for Automated Driving Using Stereo Vision and Semantic Information}. In:
  Proc. IEEE Intl. Conf. Intelligent Transportation Systems (2019)

\bibitem{bts}
Lee, J.H., Han, M.K., Ko, D.W., et~al.: From big to small: Multi-scale local
  planar guidance for monocular depth estimation (2019)

\bibitem{li2021anchor}
Li, J., Dai, H., Shao, L., Ding, Y.: Anchor-free 3d single stage detector with
  mask-guided attention for point cloud. In: Proceedings of the 29th ACM
  International Conference on Multimedia. pp. 553--562 (2021)

\bibitem{li2021voxel}
Li, J., Dai, H., Shao, L., Ding, Y.: From voxel to point: Iou-guided 3d object
  detection for point cloud with voxel-to-point decoder. In: Proceedings of the
  29th ACM International Conference on Multimedia. pp. 4622--4631 (2021)

\bibitem{li20203d}
Li, J., Luo, S., Zhu, Z., Dai, H., Krylov, A.S., Ding, Y., Shao, L.: 3d
  iou-net: Iou guided 3d object detector for point clouds. arXiv preprint
  arXiv:2004.04962  (2020)

\bibitem{li2021p2v}
Li, J., Sun, Y., Luo, S., Zhu, Z., Dai, H., Krylov, A.S., Ding, Y., Shao, L.:
  P2v-rcnn: point to voxel feature learning for 3d object detection from point
  clouds. IEEE Access  \textbf{9},  98249--98260 (2021)

\bibitem{stereorcnn}
Li, P., Chen, X., Shen, S.: Stereo r-cnn based 3d object detection for
  autonomous driving. In: CVPR (2019)

\bibitem{gfl}
Li, X., Wang, W., Wu, L., et~al.: Generalized focal loss: Learning qualified
  and distributed bounding boxes for dense object detection. In: NIPS (2020)

\bibitem{lin2017focal}
Lin, T.Y., Goyal, P., Girshick, R., He, K., Doll{\'a}r, P.: Focal loss for
  dense object detection. In: Proceedings of the IEEE international conference
  on computer vision. pp. 2980--2988 (2017)

\bibitem{lin2014microsoftcoco}
Lin, T.Y., Maire, M., Belongie, S., Hays, J., Perona, P., Ramanan, D.,
  Doll{\'a}r, P., Zitnick, C.L.: Microsoft coco: Common objects in context. In:
  European conference on computer vision. pp. 740--755. Springer (2014)

\bibitem{scnet}
Liu, J., Hou, Q., Cheng, M., et~al.: Improving convolutional networks with
  self-calibrated convolutions. In: CVPR (2020)

\bibitem{liu2022convnext}
Liu, Z., Mao, H., Wu, C.Y., et~al.: A convnet for the 2020s. arXiv preprint
  arXiv:2201.03545  (2022)

\bibitem{mimicdet}
Lu, X., Li, Q., et~al.: Mimicdet: Bridging the gap between one-stage and
  two-stage object detection. In: ECCV (2020)

\bibitem{gupnet}
Lu, Y., Ma, X., Y~ang, L., et~al.: Geometry uncertainty projection network for
  monocular 3d object detection. arXiv preprint arXiv:2107.13774  (2021)

\bibitem{luo2021m3dssd}
Luo, S., Dai, H., Shao, L., Ding, Y.: M3dssd: Monocular 3d single stage object
  detector. In: Proceedings of the IEEE/CVF Conference on Computer Vision and
  Pattern Recognition. pp. 6145--6154 (2021)

\bibitem{rethinking}
Ma, X., Liu, S., Xia, Z., et~al.: Rethinking pseudo-lidar representation. In:
  ECCV (2020)

\bibitem{AMOD}
Ma, X., Wang, Z., Li, H., et~al.: Accurate monocular 3d object detection via
  color-embedded 3d reconstruction for autonomous driving. In: ICCV (2019)

\bibitem{monodle}
Ma, X., Zhang, Y., Xu, D., et~al.: Delving into localization errors for
  monocular 3d object detection. In: CVPR (2021)

\bibitem{clocs}
Pang, S., Morris, D.D., Radha, H.: Clocs: Camera-lidar object candidates fusion
  for 3d object detection. In: {IROS} (2020)

\bibitem{dd3d}
Park, D., Ambrus, R., Guizilini, V.o.: Is pseudo-lidar needed for monocular 3d
  object detection? In: ICCV (2021)

\bibitem{LPCG}
Peng, L., Liu, F., Yu, Z., et~al.: Lidar point cloud guided monocular 3d object
  detection. CoRR  (2021)

\bibitem{fpointnet}
Qi, C.R., Wei, L., Wu, C., et~al.: Frustum pointnets for 3d object detection
  from rgb-d data. In: CVPR (2018)

\bibitem{pointnet}
Qi, C.R., Su, H., Mo, K., et~al.: Pointnet: Deep learning on point sets for 3d
  classification and segmentation. In: CVPR (2017)

\bibitem{pointnet++}
Qi, C.R., Yi, L., Su, H., et~al.: Pointnet++: Deep hierarchical feature
  learning on point sets in a metric space. In: NIPS (2017)

\bibitem{monogrnet}
Qin, Z., Wang, J., Lu, Y.: Monogrnet: A geometric reasoning network for 3d
  object localization. AAAI  (2019)

\bibitem{CADDN}
Reading, C., Harakeh, A., Chae, J., Waslander, S.L.: Categorical depth
  distribution network for monocular 3d object detection. CVPR  (2021)

\bibitem{fitnets}
Romero, A., Ballas, N., Kahou, S.E., et~al.: Fitnets: Hints for thin deep nets.
  In: ICLR (2015)

\bibitem{pvrcnn}
Shi, S., Guo, C., Jiang, L., et~al.: Pv-rcnn: Point-voxel feature set
  abstraction for 3d object detection. In: CVPR (2020)

\bibitem{pointrcnn}
Shi, S., Wang, X., Li, H.: Pointrcnn: 3d object proposal generation and
  detection from point cloud. In: CVPR (2019)

\bibitem{MonoRCNN}
Shi, X., Ye, Q., Chen, X., et~al.: Geometry-based distance decomposition for
  monocular 3d object detection. In: ICCV (2021)

\bibitem{arewemissing}
Simonelli, A., Bul{\`{o}}, S.R., Porzi, L., et~al.: Demystifying pseudo-lidar
  for monocular 3d object detection. CoRR  \textbf{abs/2012.05796} (2020)

\bibitem{disprcnn}
Sun, J., Chen, L., Xie, Y., et~al.: Disp r-cnn: Stereo 3d object detection via
  shape prior guided instance disparity estimation. In: CVPR (2020)

\bibitem{tan2019efficientnet}
Tan, M., Le, Q.: Efficientnet: Rethinking model scaling for convolutional
  neural networks. In: International conference on machine learning. pp.
  6105--6114. PMLR (2019)

\bibitem{PCT}
Wang, L., Zhang, L., Zhu, Y., et~al.: Progressive coordinate transforms for
  monocular 3d object detection. In: NIPS (2021)

\bibitem{PL}
Wang, Y., Chao, W.L., Garg, D., et~al.: Pseudo-lidar from visual depth
  estimation: Bridging the gap in 3d object detection for autonomous driving.
  In: CVPR (2019)

\bibitem{monopl}
Weng, X., Kitani, K.: {Monocular 3D Object Detection with Pseudo-LiDAR Point
  Cloud}. arXiv:1903.09847  (2019)

\bibitem{trainingshallow}
Xu, Z., Hsu, Y., et~al.: Training shallow and thin networks for acceleration
  via knowledge distillation with conditional adversarial networks. In: ICLR
  (2018)

\bibitem{second}
Yan, Y., Mao, Y., Li, B.: {SECOND:} sparsely embedded convolutional detection.
  Sensors  (2018)

\bibitem{DA-3d}
Ye, X., Du, L., Shi, Y., et~al.: Monocular 3d object detection via feature
  domain adaptation. In: ECCV (2020)

\bibitem{PL++}
You, Y., Wang, Y., Chao, W.L., et~al.: Pseudo-lidar++: Accurate depth for 3d
  object detection in autonomous driving. In: ICLR (2020)

\bibitem{monoflex}
Zhang, Y., Lu, J., Zhou, J.: Objects are different: Flexible monocular 3d
  object detection. In: CVPR (2021)

\bibitem{sessd}
Zheng, W., Tang, W., Jiang, L., et~al.: Se-ssd: Self-ensembling single-stage
  object detector from point cloud. In: CVPR (2021)

\bibitem{voxelnet}
Zhou, Y., Tuzel, O.: Voxelnet: End-to-end learning for point cloud based 3d
  object detection. CoRR  \textbf{abs/1711.06396} (2017)

\bibitem{dfrnet}
Zou, Z., Ye, X., Du, L., et~al.: The devil is in the task: Exploiting
  reciprocal appearance-localization features for monocular 3d object
  detection. In: ICCV (2021)

\end{thebibliography}

\end{document}